%% file: main.tex
\documentclass[10pt,twocolumn,letterpaper]{article}

\usepackage[pagenumbers]{cvpr}      % To produce the REVIEW version
\input{preamble}

\definecolor{cvprblue}{rgb}{0.21,0.49,0.74}
\usepackage[pagebackref,breaklinks,colorlinks,allcolors=cvprblue]{hyperref}

\def\confName{CVPR}
\def\confYear{2026}

\title{SilLang: Improving Gait Recognition with Silhouette Language Encoding}

\author{Ruiyi Zhan\textsuperscript{1}\thanks{Equal Contribution.}, Guozhen Peng\textsuperscript{1}\footnote[1]{}, Canyu Chen\textsuperscript{1}, Jian Lei\textsuperscript{2}\thanks{Corresponding Author.}, Annan Li\textsuperscript{1}\footnote[2]{}\\
{\small \textsuperscript{1}Beihang University \quad \textsuperscript{2}Tsinghua University}\\
\footnotesize{\texttt{\{zry,guozhen\_peng,chencanyu\}@buaa.edu.cn,leij23@mails.tsinghua.edu.cn,liannan@buaa.edu.cn}
}
}

\begin{document}
\maketitle
\input{sec/0_abstract}    
\input{sec/1_intro}

\input{sec/2_relatework}
\input{sec/3_method}

\input{sec/4_experiment}
\input{sec/5_conclusion}
{\small\bibliographystyle{ieeenat_fullname}\bibliography{main}}
\input{sec/X_suppl}
\end{document}

%% file: preamble.tex
\usepackage{multirow}
\usepackage{multicol}
\usepackage{booktabs}
\usepackage{bbding}
\usepackage{xcolor}

%% file: sec/0_abstract.tex
\begin{abstract}
Gait silhouettes, which can be encoded into binary gait codes, are widely adopted to representing motion patterns of pedestrian.
Recent approaches commonly leverage visual backbones to encode gait silhouettes, achieving successful performance.
However, they primarily focus on continuous visual features, overlooking the discrete nature of binary silhouettes that inherently share a discrete encoding space with natural language.
Large Language Models (LLMs) have demonstrated exceptional capability in extracting discriminative features from discrete sequences and modeling long-range dependencies, highlighting their potential to capture temporal motion patterns by identifying subtle variations.
Motivated by these observations, we explore bridging binary gait silhouettes and natural language within a binary encoding space. 
However, the encoding spaces of text tokens and binary gait silhouettes remain misaligned, primarily due to differences in token frequency and density.
To address this issue, we propose the \emph{Contour-Velocity Tokenizer}, which encodes binary gait silhouettes while reshaping their distribution to better align with the text token space.
We then establish a dual-branch framework termed \emph{Silhouette Language Model}, which enhances visual silhouettes by integrating discrete linguistic embeddings derived from LLMs.
Implemented on mainstream gait backbones, SilLang consistently improves state-of-the-art methods across SUSTech1K, GREW, and Gait3D.
\end{abstract}

%% file: sec/1_intro.tex
\section{Introduction}
\label{sec:intro}
\begin{figure}[htb]
	\centering
	\includegraphics[width=\linewidth]{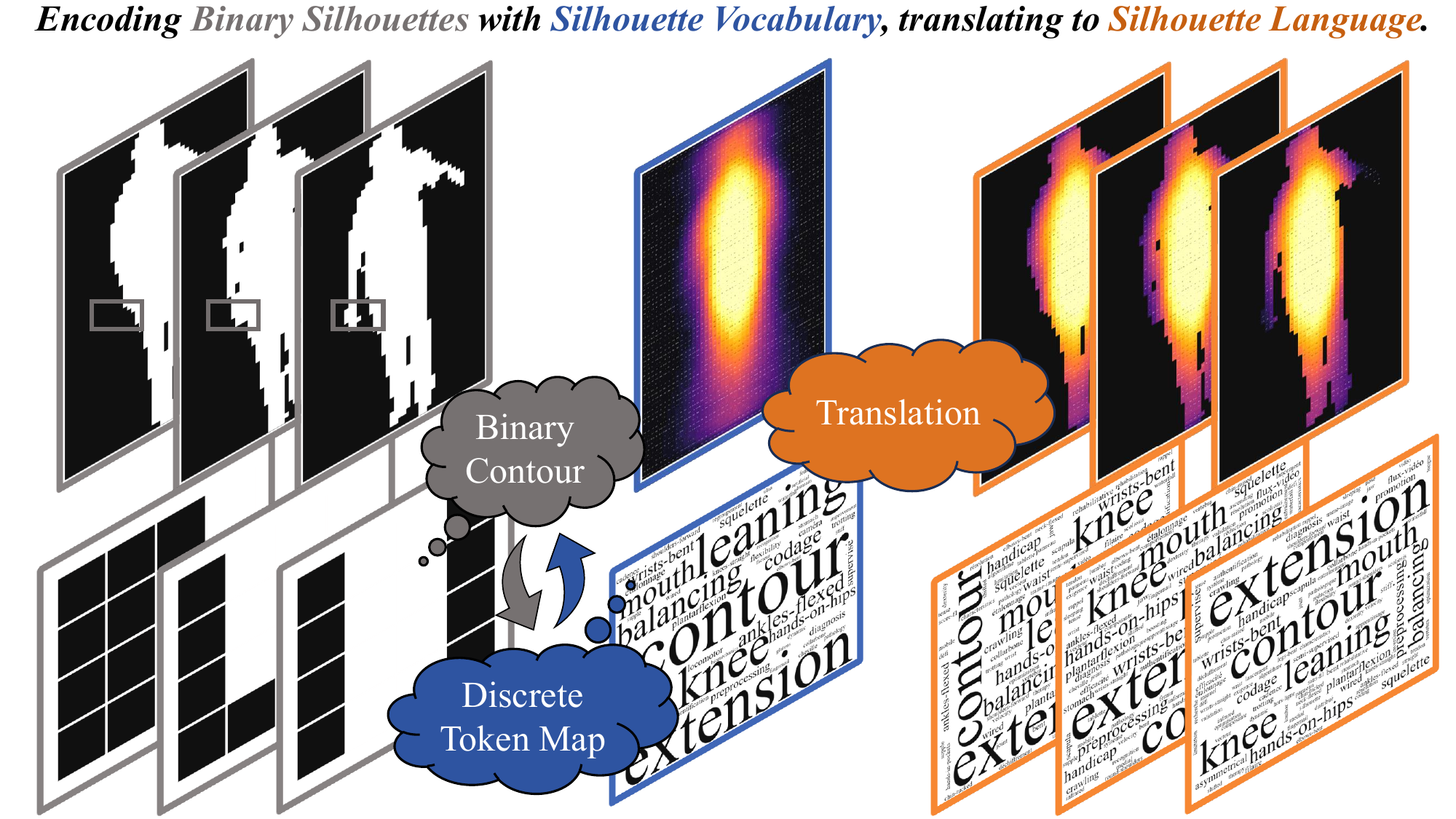}
    \caption{Binary gait silhouettes are encoded with the proposed Contour-Velocity Tokenizer, thereby forming an implicit silhouette vocabulary within the text token space. Since low-frequency silhouette tokens denote subtle walking patterns, shifts in the token frequency distribution capture fine-grained motion cues.}
	\label{figure_vis_sil_to_word}
\end{figure}
Gait recognition is a biometric technique that identifies individuals based on walking patterns. As a non-contact modality, gait can be captured remotely without subject cooperation, which makes it particularly suitable for surveillance and security applications~\cite{ref_security_application_1, ref_security_application_2, ref_security_application_3}. 

Since pedestrian gait patterns can be typically represented as binary silhouette sequences obtained through segmentation, which remove non-gait factors such as clothing variations and maintain the essential motion information, silhouette-based gait recognition has become one of the most widely adopted approaches. Recently, a number of advanced approaches~\cite{ref_waveloss, ref_gaitllm, ref_glgait, ref_deepgaitv2} have treated silhouettes as ordinary visual inputs analogous to natural images, employing vision encoders to extract continuous features. Although these methods achieve impressive performance, the continuous encoding inevitably smooths the binary input, disrupting the inherently sparse and discrete nature of gait silhouettes. This limitation indicates the potential benefit of encoding the discrete distribution of silhouettes as a valuable complement to visual encoders. To this end, we explore silhouette encoding schemes that inherently operate in a discrete space.

Notably, the text encoding space in Large Language Models (LLMs) is also sparse and discrete, revealing a natural connection between binary gait silhouettes and language. In LLMs, each word in the predefined vocabulary is represented by a word identifier (ID), typically expressed as a binary one-hot vector, implying that they share a common discrete encoding space. Motivated by this observation, we construct an implicit vocabulary for gait silhouettes, translating motion patterns into silhouette tokens that can be embedded within the LLM framework. This design enables LLMs to capture subtle silhouette encoding patterns and fine-grained structural dependencies that are frequently overlooked by visual encoders. Leveraging the discrete priors and contextual reasoning abilities of pretrained LLMs~\cite{ref_qwen3_embedding, ref_qwen3} therefore provides a promising direction for enhancing visual features of sparse binary gait silhouettes.

Nevertheless, the token density (i.e., the number of tokens required to represent one word or one silhouette) and frequency distribution in binary gait silhouettes differ substantially from those in natural language vocabularies. For instance, a typical sentence of 20 words can be represented by about 20 text tokens, whereas a gait sequence of 30 silhouettes ($64\times44$) requires over 20,000 tokens due to the high pixel redundancy, as illustrated in Figure~\ref{figure_vis_sil_to_word}. Therefore, aligning binary silhouettes with the text token space introduces two major challenges: 1) reducing the token density required to represent silhouettes. 2) reshaping the token frequency distribution to better match that of language.

To address these issues, we propose the \emph{Contour-Velocity Tokenizer} (CVT), a novel module that encodes binary gait silhouettes into the text token space. The tokenizer adjusts the intrinsic distribution of binary silhouettes to approximate the statistical properties of natural language and embeds them into a space compatible with LLMs. As shown in Figure~\ref{figure_vis_sil_to_word}, the proposed tokenizer establishes an implicit silhouette vocabulary that not only captures the binary structure but also emphasizes fine-grained motion cues, with their most similar tokens in the word vocabulary visualized. By bridging the gap between visual sparsity and linguistic discreteness, we propose a dual-branch framework termed the \emph{Silhouette Language Model} (SilLang). It first converts binary gait silhouette sequences into discrete tokens through the proposed CVT tokenizer. These silhouette tokens are then embedded by an LLM to produce text embeddings, while a lightweight visual branch simultaneously extracts complementary continuous features from the original silhouettes. Finally, a cross-modal alignment and fusion module integrates the textual and visual features into a unified embedding for gait recognition.

The main contributions are summarized as follows:
\begin{itemize}
\item \textbf{Silhouettes-to-Language Encoding.} We propose CVT tokenizer that directly translates binary silhouettes into discrete tokens, which are then embedded by LLMs.
\item \textbf{Similarity Analysis.} We provide a theoretical analysis of the structural alignment between silhouettes and natural language, showing that the silhouette encoding space can be regarded as a subset of the linguistic encoding space.
\item \textbf{Dual-Branch Framework.} We propose SilLang to enhance visual silhouette features by integrating discrete embeddings derived from LLMs. Extensive experiments validate the effectiveness and demonstrate the flexibility of the silhouette language branch, which can be seamlessly integrated with different visual gait backbones.
\end{itemize}

%% file: sec/2_relatework.tex
\section{Related Works}
\label{sec:relatework}

\subsection{Silhouette-based Gait Recognition}
Gait recognition identifies individuals through body shapes and motion patterns. To reduce interference from factors such as clothing or carried objects, most recent approaches employ gait silhouettes~\cite{ref_dygait, ref_hstl, ref_danet, ref_gaitgci, ref_gaitbase, ref_deepgaitv2, ref_qagait, ref_cltd, ref_gaitmoe, ref_vpnet, ref_glgait, ref_waveloss, ref_gaitllm, ref_uda, ref_freelunch, ref_sil_based_method_1, ref_sil_based_method_2, ref_sil_based_method_3_loss, ref_sil_based_method_4_gaitcsv} for recognition. Specifically, DyGait~\cite{ref_dygait} employs dynamic part learning to extract adaptive local features, GaitBase~\cite{ref_gaitbase} proposes an efficient backbone for large-scale deployment, and DeepGaitV2~\cite{ref_deepgaitv2} further enhances robustness and scalability through architectural expansion. Furthermore, GLGait~\cite{ref_glgait} designs a global-local temporal receptive field network, and VPNet~\cite{ref_vpnet} develops a trainable part-based prompt pool for dynamic incorporation.

In this work, we propose to encode binary gait silhouettes into text token space, enabling pre-trained Large Language Models (LLMs) to enhance visual gait features.

\subsection{Gait Recognition with Large Models}
Development of LLMs~\cite{ref_llm_gpt3, ref_llm_llama, ref_llm_deepseekr1} and Multimodal Large Language Models~\cite{ref_mllm_gemini, ref_mllm_gpt4o} (MLLMs) has introduced new paradigms for cross-modal downstream tasks. CLIP~\cite{ref_clip} achieves image-text feature alignment through contrastive learning, while LLaVA~\cite{ref_llava} extends this capability by integrating vision and text features through cross-modal attention mechanisms. Qwen3~\cite{ref_qwen3} further achieves state-of-the-art results across diverse benchmarks, including reasoning, coding, and multilingual understanding.
DeepSeek-OCR~\cite{ref_deepseekOCR} proposes a unified architecture that reformulates text understanding as a visual modeling task for structured document images.

Recently, large models are applied to gait recognition methods. Specifically, BigGait~\cite{ref_biggait} and BiggerGait~\cite{ref_biggergait} extract gait features by large vision models originally designed for general object detection. GaitLLM~\cite{ref_gaitllm} feeds LLMs with the features extracted from visual gait backbone. However, these frameworks neglect the intrinsic binary attribute of silhouettes, encoding them solely with general vision encoders for subsequent processing. In contrast, we propose CVT to encode binary gait silhouettes into discrete tokens, and directly enhance them by LLMs.

%% file: sec/3_method.tex
\section{Method}
\label{sec:method}

% Here is the method section.

\subsection{Translate Silhouette into Words}
\begin{figure}[t]
	\centering
	\includegraphics[width=\linewidth]{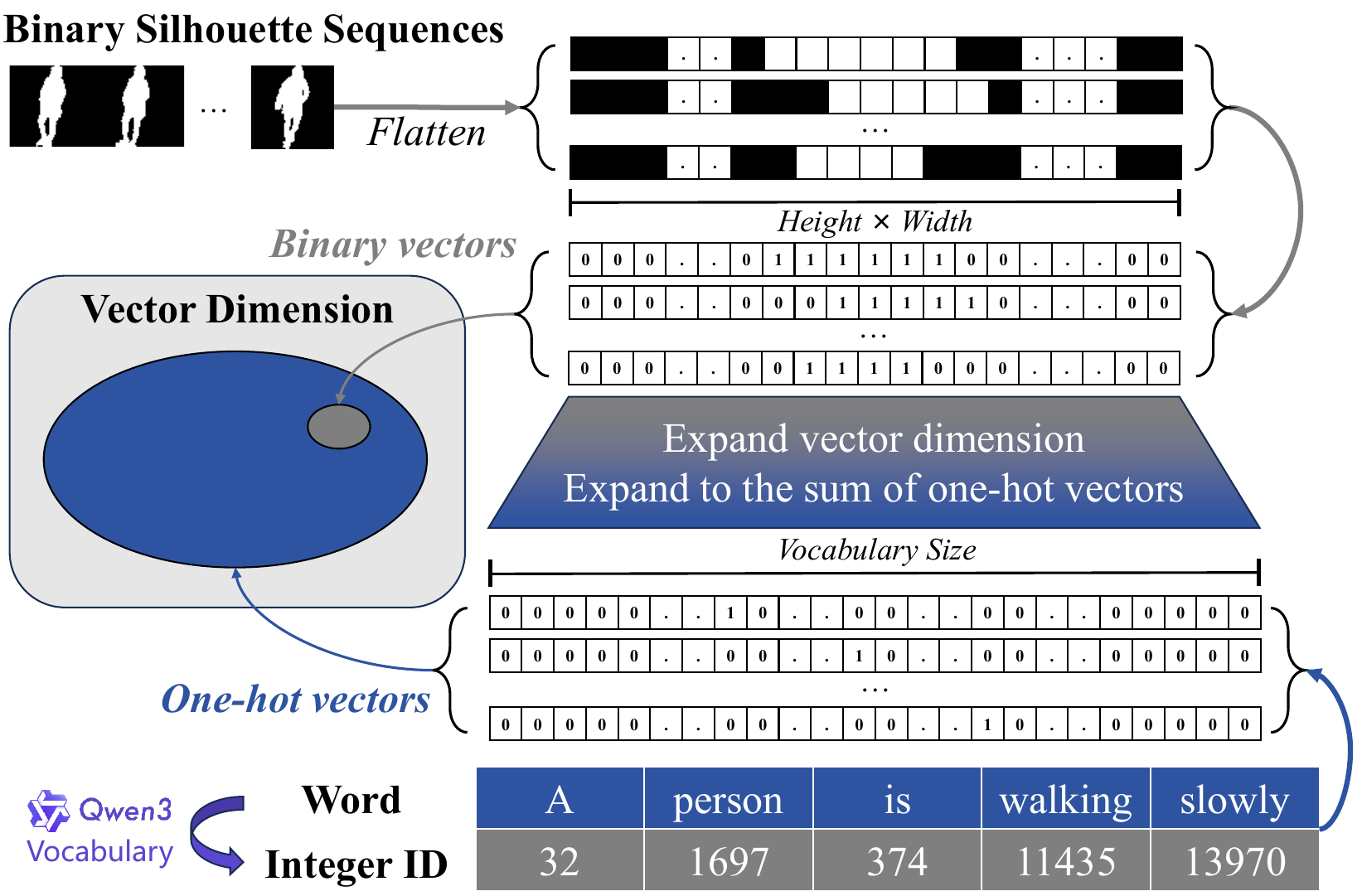}
	\caption{Aligning and translating binary gait silhouettes into natural language within a shared binary encoding space.}
	\label{figure_relationship_sil_text}
\end{figure}

In gait recognition, the walking pattern is usually represented by the silhouettes sequence of pedestrian, which is actually binary image of body contour. The typical size is $64\times44$ (e.g., in SUSTech1K~\cite{ref_sustech1k}, GREW~\cite{ref_grew} and Gait3D~\cite{ref_gait3d}).
Notably, if a binary silhouette is flattened into a one-dimensional vector (denoted as $\boldsymbol{s}$), it manifests as a binary code with a bit length of $S_L=height\times width$.
Since only the contour pixels are meaningful, the binary code is rather sparse which differs substantially from the dense distribution of gray scales of color image.  
Meanwhile, language models typically employ tokenizers and vocabularies to encode words, mapping words to unsigned integers.
Then, each word can be encoded into a one-hot vector with a length equal to the size of the vocabulary ($N$). The one-hot vector $\boldsymbol{z}^{k}$ for $k\cdot th$ word in the vocabulary can be encoded as:
\begin{equation}
\boldsymbol{z}^k=[z_1^{k},z_2^{k},\cdots,z_N^{k}]^T,
\end{equation}
where $k \in \{1,2,\dots,N\}$. And only when $i=k$ does $z_i^{k}=1$; in all other cases $z_i^{k}=0$.
Accordingly, as shown in Figure~\ref{figure_relationship_sil_text}, it can be hypothesized that \emph{the distribution of silhouettes aligns closely with that of language}.
To verify the aforementioned hypothesis, we decompose the vector of flattened silhouette $\boldsymbol{s}$ into the sum of $\boldsymbol{s}^{k}$, which is similar to $\boldsymbol{z}^k$ as follows:
\begin{equation}
\boldsymbol{s}^{k}=[s_1^{k},s_2^{k},\cdots,s_{S_L}^{k}]^T,
\end{equation}
\begin{equation}
\label{formula_decompose_silhouette_to_vector}
\boldsymbol{s}=\sum_{\boldsymbol{s}_{i}=1}\boldsymbol{s}^{i}\quad \text{where }i\leq S_L,
\end{equation}
which indicates that the encoding space of silhouette is similar to that of language.
Therefore, $\boldsymbol{s}$ can be represented by an unordered set of words when expanding $s^{k}$ to $z^{k}$ with a learnable binary matrix $\boldsymbol{B}_{N\times S_L}$:
\begin{align}
\label{formula_expand_sil_vector_to_word_vector}
\boldsymbol{Enc}(\boldsymbol{s})&=\sum_{\boldsymbol{s}_{i}=1}\boldsymbol{B}_{N\times S_L}\boldsymbol{s}^{i}\quad \text{where }i\leq S_L
\\&=\sum_{\boldsymbol{s}_{i}=1}z^{i}\quad \text{where }i\leq N,
\end{align}
where $\boldsymbol{Enc}$ encodes $\boldsymbol{s}$ into a text token space.
Specifically, binary gait silhouettes and natural language can be converted into each other when the vocabulary size $N$ equals the number of pixels $S_L$.
In the context of silhouette-based gait recognition, $S_L$ typically takes a value of 2816 ($64\times 44$).
However, for instance, $N$ used in Qwen3-Embedding~\cite{ref_qwen3_embedding} is 151,642, which is far larger than that of $S_L$. 
Then, under the condition $S_L < N$, we have $Rank(\boldsymbol{B}_{N\times S_L}) = S_L$, which implies that the silhouette encoding space is contained within the language encoding space; consequently, binary gait silhouettes can be encoded into text token space.
Furthermore, incorporating temporal information allows the binary silhouettes to be represented as $f$ ordered aggregated word vectors:
\begin{equation}
\label{formula_translate_sils_to_words}
    \{\boldsymbol{s}(t)\}=\{\sum_{\boldsymbol{s}(t)_{i}=1}z^{i}\}\quad \text{where }t=1,2,\cdots,f,
\end{equation}
where $f$ and $\boldsymbol{s}(t)$ refer to the number of frames and the $t\cdot th$ encoded silhouette $\boldsymbol{Enc}(\boldsymbol{s})$ of the sequence, respectively.
%
% %
\begin{figure*}[tb]
	\centering
	\includegraphics[width=\linewidth]{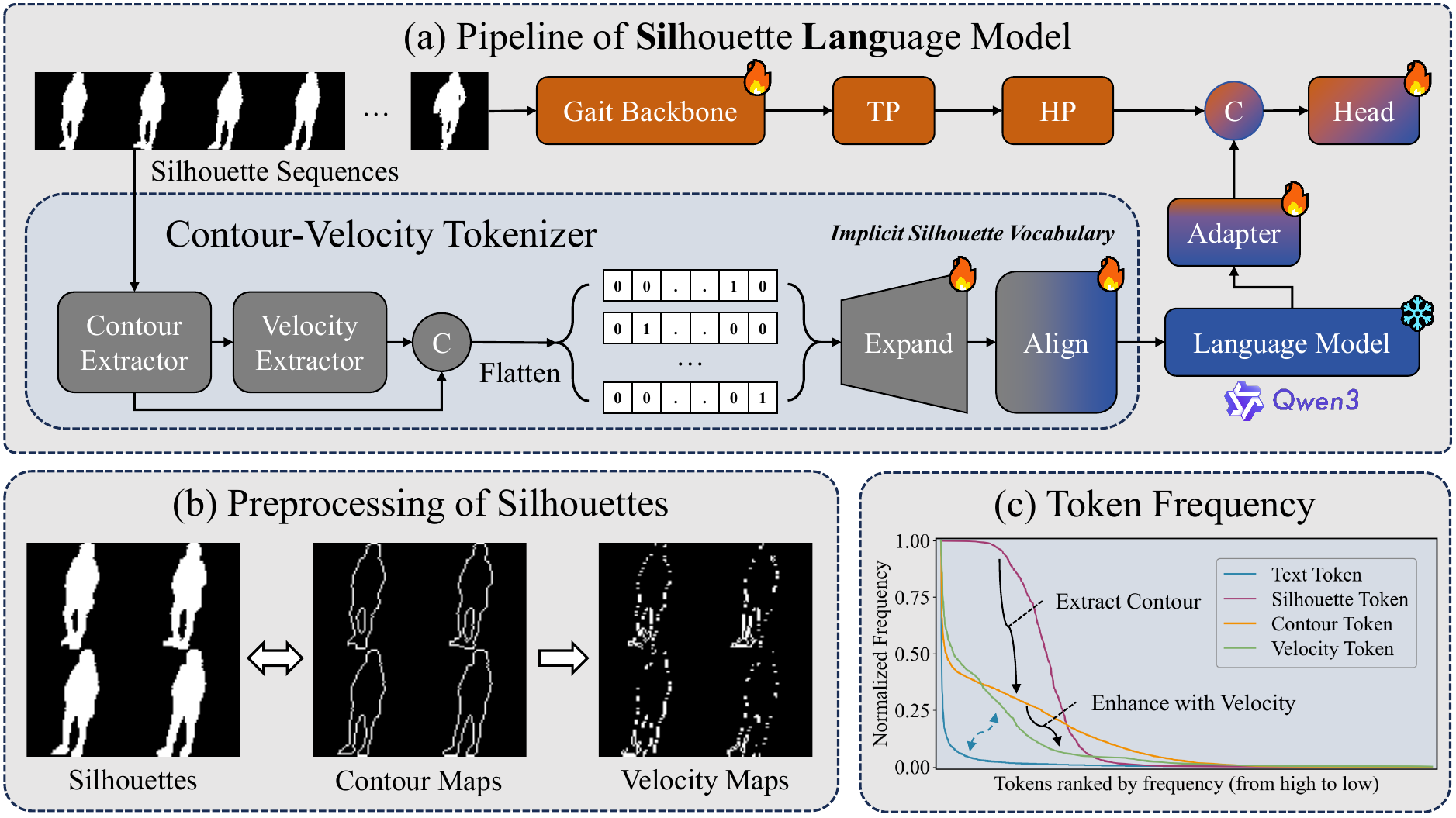}
	\caption{Pipeline of the silhouette language model. (a) The silhouette sequences are encoded with a Contour-Velocity Tokenizer for silhouette language branch, and the Adapter adapts the text embedding ($\boldsymbol{emb}_t$) to the visual embedding ($\boldsymbol{emb}_v$). In the visual branch, it keeps the same structure as other silhouette-based gait recognition model~\cite{ref_deepgaitv2, ref_glgait}. (b) Extract contour map $\boldsymbol{c}(t)$ and velocity map $\boldsymbol{v}(t)$ from silhouette $\boldsymbol{s}(t)$. (c) Visualization of the normalized token frequencies in silhouette and text vocabulary when align silhouette to language.}
	\label{figure_pipeline}
\end{figure*}
\subsection{Contour-Velocity Tokenizer for Silhouette}
 \begin{table}[t]
    \centering
    \caption{Analysis of token density in silhouettes, with 1,000 silhouette ($64\times44$) sequences sampled randomly from SUSTech1K~\cite{ref_sustech1k}, GREW~\cite{ref_grew} and Gait3D~\cite{ref_gait3d}. Token density $P_s(1)$, $P_c(1)$, and $P_v(1)$ denote the probabilities of a pixel being 1 in the silhouette, contour map, and velocity map, respectively, approximated by their observed frequencies. ACR represents the Average Compression Rate relative to the original token density.
    }
    \label{table_composition_rate_of_01}

    \begin{tabular}{c|c|c|c|c}
         \toprule
         \textbf{Density} 
         & \textbf{SUSTech1K}
         & \textbf{GREW}
         & \textbf{Gait3D}
         & \textbf{ACR}
         \\
        \midrule
        $P_s(1)$ & 21.2\% & 25.0\% & 20.2\% & 100\%\\
        $P_c(1)$ & 4.5\% & 4.1\% & 4.1\% &  19.3\%\\
        $P_v(1)$ & 1.8\% & 2.0\% & 2.1\% &  9.0\%\\
        \bottomrule
    \end{tabular}
\end{table}
\textbf{Token Density and Frequency.} While Equation~\eqref{formula_translate_sils_to_words} successfully expands $\boldsymbol{s}$ into word vectors, Table~\ref{table_composition_rate_of_01} shows that the resulting set typically contains more than 500 tokens ($P_s(1)\times S_L$).
Such an unusually high number of aggregated words is uncommon in language models, for which we introduce the concept of \emph{token density} (denoted as $P(1)$) to describe tokens in one silhouette.
Furthermore, silhouette sequences containing up to 720 frames (the typical maximum used for inference, approximately 360K tokens) substantially increase the token cost, which may exceed the maximum token capacity of the language model (e.g., 32K for Qwen3-Embedding~\cite{ref_qwen3_embedding}) and is thereby constrained by its ability to process long textual inputs.
In addition to token density, we introduce \emph{token frequency} to describe the occurrence frequency of different tokens.
As shown in Figure~\ref{figure_pipeline} (c), despite differences in token density, the token frequency distributions of text tokens ($\boldsymbol{z}^{k}$) and silhouette tokens ($\boldsymbol{s}^{k}$) remain distinct.
This misalignment in token frequency distribution weakens the ability of language model to capture distinctive motion cues encoded by low-frequency silhouette tokens.
To address both issues, we design a \emph{Contour-Velocity Tokenizer} for silhouettes, which compresses token density via map extractors and aligns the frequency distribution of silhouette tokens with that of text tokens, thereby constructing an implicit silhouette vocabulary and effectively encoding binary silhouettes into text token space.
\textbf{Contour Extractor.} Since $\boldsymbol{s}(t)$ in Equation~\eqref{formula_translate_sils_to_words} reveals that the number of $\boldsymbol{z}^{i}$ depends on the number of one-hot vectors, where $\boldsymbol{s}(t)_{i}=1$.
Thus, reducing the token density is equivalent to decreasing the number of TRUE values (white pixels).
Considering that the TRUE values within body regions are not that informative in distinguishing subtle walking patterns, we extract the contour map $\boldsymbol{c}(t)$ from $\boldsymbol{s}(t)$ by inverting the interior pixel values to 0 while preserving only the boundary pixels as 1.
Results in Table~\ref{table_composition_rate_of_01} indicate that token density $P_c(1)$ in contour map have been successfully compressed to 19.3\% of $P_s(1)$.
Furthermore, under the specific data structure of binary pedestrian silhouettes, $\boldsymbol{c}(t)$ and $\boldsymbol{s}(t)$ can be converted into each other losslessly, which is illustrated in Figure~\ref{figure_pipeline} (b).

As can be seen, the Contour-Velocity Tokenizer parses $\boldsymbol{c}(t)$ extracted from $\boldsymbol{s}(t)$, without introducing additional information loss.
The token frequency distributions of silhouette tokens, contour tokens, and text tokens are compared in Figure~\ref{figure_pipeline}.
It demonstrates that the frequency distribution of silhouette tokens differs significantly from that of text tokens, whereas the proposed contour extractor effectively shifts it toward that of text tokens.

\textbf{Velocity Extractor.} Although the frequency distribution of contour tokens is closer to that of text tokens, a notable disparity persists, especially within the low-to-mid frequency range.
To further address this issue, we extract the differences between frames of $\boldsymbol{c}(t)$ to construct the velocity map $\boldsymbol{v}(t)$.
Compared with contour tokens, velocity tokens exhibit a distribution more closely aligned with that of text tokens in the low-to-mid frequency range.
This alignment facilitates a clearer distinction between different silhouettes, as it becomes less likely for velocity maps extracted from distinct silhouettes to share the same velocity token.
Moreover, as shown in Figure~\ref{figure_pipeline} (b) and \ref{figure_pipeline} (c), the velocity map can capture motion as temporal features, thereby enriching the encoded tokens with gait-relevant information.
Additionally, as presented in Table~\ref{table_composition_rate_of_01}, velocity tokens exhibit a lower token density (compressed to 9.0\%) compared to contour tokens, further narrowing the gap between silhouettes and language.
\textbf{Expand and Align.} After extracting the contour map and velocity map, we encode silhouettes with the contour and velocity tokens, and employ a Multi-Layer Perceptron (MLP) to expand and align the encoded silhouettes with the text token space. 
Since the token frequency within the contour and velocity maps still shows a slight misalignment with that of text tokens, we introduce a learnable coefficient in the align process to weight each token, thereby balancing the token frequency distribution. 
The coefficient is initialized as the reciprocal of the frequency of each token and is normalized using the contour frequency, which provides a balanced scaling between static body shape (silhouette tokens) and dynamic motion cues (velocity tokens). The coefficients are estimated from the training set as prior knowledge.
Finally, the process is described in Equation~\eqref{formula_tokenizer}.
\begin{equation}
\label{formula_tokenizer}
    \boldsymbol{token}_{sil}(t)=EA(Concat(\boldsymbol{c}(t),\boldsymbol{v}(t))),
\end{equation}
where $EA$ expands the contour and velocity tokens into the text token space and aligns their frequency distribution with that of text tokens.
The resulting $\{\boldsymbol{token}_{sil}(t)|t=1,2,\cdots f\}$ are then used as inputs to the language model.
\begin{table}[t]
    \centering
    \caption{Recognition results of different embedding fusion methods on Gait3D with DeepGaitV2-P3D.
    }
    \label{table_token_fusion_exploration}

    \begin{tabular}{c|c|c|c}
         \toprule
         \textbf{Fusion method} 
         & \textbf{Rank-1}
         & \textbf{Rank-5}
         & \textbf{mAP}
         \\
        \midrule
        Attention & \underline{73.3} & 85.7 & \underline{64.3} \\
        Token Concat & 71.4 & \underline{86.0} & 63.8 \\
        Channel Concat & \textbf{75.2} & \textbf{87.2} & \textbf{67.1} \\
        \bottomrule
    \end{tabular}
\end{table}
\subsection{Exploration on Silhouette Language Model}
Since the silhouette sequences in gait datasets (e.g., 18K in Gait3D~\cite{ref_gait3d}) are much less than that used for LLM training, while Qwen3-Embedding-0.6B~\cite{ref_qwen3_embedding} is trained on approximately 150 million pairs of synthetic data.
And the silhouettes are of low quality~\cite{ref_qagait, ref_rsanet, ref_gqan}, which lead to ambiguous $\boldsymbol{c}(t)$ and $\boldsymbol{v}(t)$, amplifies the noise introduced during the silhouettes extraction process.
We propose to frozen the pre-trained LLMs instead of training or finetuning them.
Therefore, as illustrated in Figure~\ref{figure_pipeline} (a), we adopted a dual-branch architecture for gait recognition.
The visual branch follows the design paradigm of mainstream gait recognition models (e.g., DeepGaitV2~\cite{ref_deepgaitv2}, GLGait~\cite{ref_glgait}) and consists of four main components~\cite{ref_gaitbase}: Gait Backbone, Temporal Pooling (TP), Horizontal Pooling (HP), and the recognition Head (Head)~\cite{ref_BNNnecks}. 
The other one is the silhouette language branch, which enhances the visual feature $\boldsymbol{emb_{v}}$ obtained from HP module.
As shown in Equation~\eqref{formula_emb_eval}, $\boldsymbol{emb_{eval}}$ for recognition head fusions $\boldsymbol{emb_{t}}$ and $\boldsymbol{emb_{v}}$ from language model and gait backbone, respectively.
\begin{equation}
\label{formula_emb_eval}
\boldsymbol{emb_{eval}}=Concat(\boldsymbol{emb_{v}},Adapter(\boldsymbol{emb_{t}})),
\end{equation}
where $Adapter$ is an MLP that adapts $\boldsymbol{emb_{t}}$ to gait recognition task.
Furthermore, since the model is not jointly trained on a sufficiently large and diverse labeled datasets of $\boldsymbol{s}(t)$–$\boldsymbol{token}_{sil}(t)$-$text$ pairs, the embedding spaces of $\boldsymbol{emb_{v}}$ and $\boldsymbol{emb_{t}}$ are not fully aligned.
Consequently, as shown in Table~\ref{table_token_fusion_exploration}, concatenation along the channel dimension yields better performance than attention-based fusion.

%% file: sec/4_experiment.tex
\section{Experiment}
\label{sec:experiment}

\begin{table*}[tb]
    \centering
    \caption{Comparison of Rank-1, 5 accuracy and mean Average Precision (\%) on Gait3D and GREW, where *-P3D and *-GL refer to the gait backbone, DeepGaitV2-P3D and GLGait, respectively. The superscript $\dagger$ denotes results obtained using 128 channels, whereas our experiments are conducted with 64 channels due to limited computational resources.}
    \label{table_main_Comparison}
    \setlength{\tabcolsep}{2mm}{

    % \begin{tabular*}{\linewidth}{c|c|ccc|ccc}
    \begin{tabular}{c|c|ccc|cc}
        \toprule
        \multirow{2}{*}{\textbf{Method}} & \multirow{2}{*}{\textbf{Publication}}
         & \multicolumn{3}{c|}{\textbf{Gait3D}} & \multicolumn{2}{c}{\textbf{GREW}}  \\
         % \midrule{3-8}
        & & Rank-1 & Rank-5 & mAP & Rank-1 & Rank-5  \\
        \midrule
        GaitSet~\cite{ref_gaitset} & AAAI 2019 & 36.7 & 58.3 & 30.0 & 46.3 & 63.6   \\
        GaitPart~\cite{ref_gaitpart} & CVPR 2020 & 28.2 & 47.6 & 21.6 & 44.0 & 60.7  \\
        GaitGL~\cite{ref_gaitgl} & ICCV 2021 & 29.7 & 48.5 & 22.3 & 47.3 & 63.6 \\ 
        SMPLGait~\cite{ref_gait3d} & CVPR 2022 & 53.2 & 71.0 & 42.4 & - & - \\
        DyGait~\cite{ref_dygait} & ICCV 2023 & 66.3 & 80.8 & 56.4 & 71.4 & 83.2 \\
        HSTL~\cite{ref_hstl} & ICCV 2023 & 61.3 & 76.3 & 55.5 & 62.7 & 76.6 \\
        GaitGCI~\cite{ref_gaitgci} & CVPR 2023 & 57.2 & 74.5 & 45.0 & 68.5 & 80.8 \\
        GaitBase~\cite{ref_gaitbase} & CVPR 2023 & 64.6 & 79.6 & 55.5 & 60.1 & 74.5  \\
        QAGait~\cite{ref_qagait} & AAAI 2024 & 67.0 & 81.5 & 56.5 & 59.1 & -  \\
        CLTD~\cite{ref_cltd} & ECCV 2024 & 69.7 & 85.2 & - & 78.0 & 87.8  \\
        GaitMoE~\cite{ref_gaitmoe} & ECCV 2024 & 73.7 & - & 66.2 & 79.6 & 89.1  \\
        VPNet~\cite{ref_vpnet} & CVPR 2024 & 75.4 & 87.1 & - & 80.0 & 89.4  \\
         GLGait~\cite{ref_glgait} & MM 2024 & \underline{77.6} (77.7$^\dagger$) & \textbf{88.4} (88.9$^\dagger$) & \underline{69.6} (70.6$^\dagger$) & 80.0 (82.8$^\dagger$) & 89.4 (91.1$^\dagger$)  \\
         WaveLoss~\cite{ref_waveloss}  & AAAI 2025 & 75.6 & 88.4 & 66.5 & - & - \\
         DeepGaitV2-P3D~\cite{ref_deepgaitv2} & TPAMI 2025 & 74.4 (75.0$^\dagger$) & 88.0 & 65.8 & 77.7 & 87.9  \\
         GaitLLM-P3D~\cite{ref_gaitllm}  & CVPR 2025 & 76.5 & 88.1 & 68.3 & 79.8 & 89.5 \\
          \midrule
         SilLang-P3D & \multirow{2}{*}{Ours} & 76.9 & \underline{88.2} & 68.8 & \underline{80.6} & \underline{89.9}  \\
         SilLang-GL &  & \textbf{78.4} & 87.8 & \textbf{70.3} & \textbf{81.2} & \textbf{90.1}  \\
        \bottomrule
    \end{tabular}
    }
\end{table*}
\begin{table*}[tb]
    \centering
    \caption{Results with different attributes on SUSTech1K, where $*$ indicates that the method is reproduced with the same configuration.}
    \label{table_sustech1k_results}
    \begin{tabular}{c|cccccccc|c}
        \toprule
        \multirow{2}{*}{\textbf{Method}} 
        & \multicolumn{8}{c|}{\textbf{Probe Sequence (Rank-1)}}
        & \multicolumn{1}{c}{\textbf{Overall}}  \\
        & Normal & Bag & Clothing & Carrying & Umbrella & Uniform & Occlusion & Night & Rank-1    \\
        \midrule
        GaitSet~\cite{ref_gaitset} & 69.1 & 68.3 & 37.4 & 65.0 & 63.1 & 61.0 & 67.2 & 23.0 & 65.0   \\
        GaitPart~\cite{ref_gaitpart} & 62.2 & 62.8 & 33.1 & 59.5 & 57.3 & 54.9 & 57.2 & 21.8 & 59.2  \\
        GaitGL~\cite{ref_gaitgl} & 67.1 & 66.2 & 35.9 & 63.3 & 61.6 & 58.1 & 66.6 & 17.9 & 63.1  \\
        GaitBase~\cite{ref_gaitbase} & 81.5 & 77.5 & 49.6 & 75.8 & 75.6 & 76.7 & 81.4 & 25.9 & 76.1  \\
        DeepGaitV2-P3D$^*$ \cite{ref_deepgaitv2} & \textbf{86.9} & 82.4 & 48.9 & 79.2 & 83.3 & 81.8 & 85.0 & 27.7 & 80.1 \\
        GaitLLM-P3D$^*$ \cite{ref_gaitllm} & \underline{86.4} & \underline{83.1} & \underline{53.1} & \underline{80.5} & \underline{85.3} & \underline{83.9} & \underline{88.0} & \underline{27.7} & \underline{81.6} \\
        \midrule
        SilLang-P3D (Ours) & 86.3 & \textbf{83.2} & \textbf{53.6} & \textbf{81.1} & \textbf{85.7} & \textbf{85.0} & \textbf{88.1} & \textbf{28.3} & \textbf{82.0}  \\
        \bottomrule
    \end{tabular}
\end{table*}
\subsection{Datasets and Implementation Details}
\textbf{Gait3D}~\cite{ref_gait3d} comprises 4,000 subjects, 25,309 sequences, and 3,279,239 frames. Specifically, 3,000 subjects are partitioned for training, while the remaining 1,000 subjects constitute the test set. During the evaluation phase, one sequence from each subject is utilized as the probe, and all remaining sequences form the gallery set.
\\
\textbf{GREW}~\cite{ref_grew} represents one of the largest gait recognition datasets collected in unconstrained real-world environments. The dataset was acquired using 882 cameras distributed across extensive public areas, amounting to approximately 3,500 hours of $1,920\times 1,080$ high-resolution streams. In total, it contains 26,345 subjects and 128,671 sequences, which are partitioned into training and test sets comprising 20,000 and 6,000 subjects, respectively.
\\
\textbf{SUSTech1K}~\cite{ref_sustech1k} is a laboratory-controlled gait dataset acquired using LiDAR sensors and RGB cameras. It comprises 25,239 sequences from 1,050 subjects and covers a wide range of variations, including visibility, views, occlusions, clothing, carrying, and scenes.
\\
\textbf{Implementation Details.} In our experiments, we employ Qwen3-Embedding-0.6B~\cite{ref_qwen3_embedding} as the language model for the silhouette language branch. For the visual branch, we adopt DeepGaitV2-P3D~\cite{ref_deepgaitv2} and GLGait~\cite{ref_glgait} as the backbones. For Gait3D and GREW, DeepGaitV2-P3D is configured with a 22-layer architecture, while GLGait uses the large model (GLGait-L~\cite{ref_glgait}). Since SUSTech1K contains only 250 identities in the training set, a 10-layer configuration of DeepGaitV2-P3D is adopted to avoid overfitting.

In the training stage, the input silhouettes are resized to $64\times 44$, which is simultaneously fed into two parallel branches. The optimizer is Stochastic Gradient Descent (SGD), while the weight decay and the momentum are set to 0.0005 and 0.9. And the parameters of language model are frozen, while other parameters remain trainable. CTL~\cite{ref_glgait} from GLGait is used for Gait3D and GREW, while triplet~\cite{ref_tripletloss} and cross-entropy~\cite{ref_crossentropyloss} loss are used for SUSTech1K. Then, we train the model with a batch size of $32\times 4$ (4 sequences for each pedestrian, 32 pedestrians in total) on Gait3D and GREW, while $8\times 8$ on SUSTech1K. 1) On Gait3D, the training iteration is 120k. The learning rate starts at 0.1 and is subsequently decreased by a factor of 0.1 at iterations (40k, 80k, 100k). 2) On GREW, the training iteration is 180k. The learning rate starts at 0.05 and is subsequently decreased by a factor of 0.2 at iterations (60k, 120k, 150k). 3) On SUSTech1K, the training iteration is 50k. The learning rate starts at 0.1 and is subsequently decreased by a factor of 0.1 at iterations (20k, 30k, 40k).

\subsection{Performance Comparison}
We integrate SilLang with silhouette-based gait backbones, including DeepGaitV2-P3D~\cite{ref_deepgaitv2} and GLGait~\cite{ref_glgait}, which are denoted as SilLang-P3D and SilLang-GL, respectively. This section presents a comparison between SilLang and other state-of-the-art gait recognition methods on the Gait3D, GREW, and SUSTech1K datasets, evaluated by Rank-1, 5 accuracy and mean Average Precision (mAP).

\textbf{Improvement on Metrics.} As shown in Table~\ref{table_main_Comparison}, SilLang achieves consistent improvements across multiple gait silhouette datasets. Specifically, our best results reach 78.4\% and 81.2\% in Rank-1 accuracy on the Gait3D and GREW datasets, respectively. When integrated with the DeepGaitV2-P3D, SilLang-P3D improves Rank-1 accuracy and mAP on Gait3D by +2.5\% and +3.0\%, and increases Rank-1 accuracy on GREW by +2.9\%. 

Furthermore, as shown in Table~\ref{table_sustech1k_results}, when applied to the SUSTech1K dataset, SilLang-P3D further enhances model robustness to diverse variations, and increases the overall Rank-1 accuracy by +1.9\%. These results demonstrate that the language-encoded silhouettes effectively enhances the discriminative capability of the model across diverse and challenging scenarios.

\textbf{Effectiveness on Different Backbones.} SilLang also exhibits strong generalization capability across various backbone architectures, including both 3D CNN-based~\cite{ref_cnn_for_p3d_1, ref_cnn_for_p3d_2} DeepGaitV2-P3D and Transformer-based~\cite{ref_attn_for_glgait} GLGait. As shown in Table~\ref{table_main_Comparison}, the consistent improvements observed across different architectures indicate that the silhouette language branch provides complementary information to the visual branch. Specifically, while the visual gait backbone smooths continuous pixel variations and shows limited sensitivity to fine contour details, the proposed tokenizer amplifies these discrete motion cues, enabling the LLM to embed them more distinctly and complement the visual features.

Moreover, since the silhouette language is derived directly from binary gait silhouettes, SilLang introduces no dependency on external data sources, thereby avoiding the limitations and quality constraints typically associated with additional modalities.
\begin{table}[t]
    \centering
    \caption{Ablation study on components of the Contour-Velocity Tokenizer, including the velocity extractor ($\boldsymbol{v}(t)$) and dimension expansion and align module ($EA$).
    }
    \label{table_method_ablation_study}

    \begin{tabular}{c|c|c|c|c}
         \toprule
         $\boldsymbol{v}(t)$
         &
         \textbf{$EA$}
         & \textbf{Rank-1}
         & \textbf{Rank-5}
         & \textbf{mAP}
         \\
        \midrule
        \XSolidBrush & \XSolidBrush & 73.7 & 87.1 & 64.9 \\
        \XSolidBrush & \CheckmarkBold & 74.0 & \textbf{88.1} & \underline{66.2} \\
        \CheckmarkBold & \XSolidBrush & \underline{74.5} & 86.9 & \underline{66.2} \\
        \CheckmarkBold & \CheckmarkBold & \textbf{75.2} & \underline{87.2} & \textbf{67.1} \\
        \bottomrule
    \end{tabular}
\end{table}
\begin{table}[t]
    \centering
    \caption{Ablation study on branches selected for training, where the Backbone and LLM refer to the gait backbone (DeepGaitV2-P3D) and the large language model (Qwen3-Embedding-0.6B).
    }
    \label{table_training_modules_ablation_study}

    \begin{tabular}{c|c|c|c|c}
         \toprule
         \textbf{Backbone}
         &
         \textbf{LLM}
         & \textbf{Rank-1}
         & \textbf{Rank-5}
         & \textbf{mAP}
         \\
        \midrule
        \XSolidBrush & \XSolidBrush & 71.6 & \underline{86.8} & 64.4 \\
        \XSolidBrush & \CheckmarkBold & 72.1 & 86.1 & 64.0 \\
        \CheckmarkBold & \XSolidBrush & \textbf{75.2} & \textbf{87.2} & \textbf{67.1} \\
        \CheckmarkBold & \CheckmarkBold & \underline{73.5} & \textbf{87.2} & \underline{65.4} \\
        \bottomrule
    \end{tabular}
\end{table}
\begin{table}[t]
    \centering
    \caption{Ablation study on silhouette vocabulary size $S_L$ (pixels), where - is the reproduced baseline without language model.
    }
    \label{table_pixels_ablation_study}

    \begin{tabular}{c|c|c|c}
         \toprule
         \textbf{$S_L$ ($Height\times Width$)}
         & \textbf{Rank-1}
         & \textbf{Rank-5}
         & \textbf{mAP}
         \\
        \midrule
        - & 74.2 &	\underline{86.9} & \textbf{67.1} \\
         704 ($32\times22$)  & \underline{74.7} & 86.6 & \underline{65.6} \\
        2816 ($64\times44$)  & \textbf{75.2} & \textbf{87.2} & \textbf{67.1} \\
        \bottomrule
    \end{tabular}
\end{table}
\begin{table}[t]
    \centering
    \caption{Ablation study on different LLM size. 
    }
    \label{table_llm_model_ablation_study}

    \begin{tabular}{c|c|c|c}
         \toprule
         \textbf{Language Model}
         & \textbf{Rank-1}
         & \textbf{Rank-5}
         & \textbf{mAP}
         \\
        \midrule
        Qwen-Embedding-0.6B & \underline{76.9} & \underline{88.2} & \underline{68.8} \\
        Qwen-Embedding-4B  & \textbf{77.2} & \textbf{88.8} & \textbf{69.3} \\
        \bottomrule
    \end{tabular}
\end{table}
\subsection{Ablation Study}
All ablation studies are conducted on the Gait3D dataset using SilLang-P3D. The experiments reported in Table~\ref{table_method_ablation_study}, Table~\ref{table_training_modules_ablation_study}, and Table~\ref{table_pixels_ablation_study} are trained for 60K iterations to reduce computational overhead and training time. In contrast, the results in Table~\ref{table_llm_model_ablation_study} are obtained after 120K iterations to further demonstrate the embedding capability of larger LLMs.

\textbf{Effect of the Contour-Velocity Tokenizer.} The effectiveness of proposed tokenizer is shown in Table~\ref{table_method_ablation_study}. As illustrated in Figure~\ref{figure_pipeline} (b), since $\boldsymbol{s}(t)$ and $\boldsymbol{c}(t)$ are equivalent representations, we focus on $\boldsymbol{v}(t)$ and the $EA$ module.

The results indicate that both components contribute positively to recognition accuracy. The velocity map $\boldsymbol{v}(t)$ enriches fine-grained motion representations by incorporating temporal cues, whereas the $EA$ module further aligns the token frequency distribution. This expand and align process essentially constructs an implicit silhouette vocabulary that maps contour and velocity tokens into the text token space, thereby enabling more effective exploitation of the contextual embedding capability of LLMs.

\textbf{Training Branch Configuration.} Table~\ref{table_training_modules_ablation_study} compares different configurations for freezing and training the visual gait backbone and the LLM. The best performance is achieved when the LLM is frozen and only the gait backbone is trained. This configuration offers two main advantages. First, the silhouette dataset contains only about 18K\cite{ref_gait3d} sequences for the silhouette language branch, which may cause overfitting if the LLM is trained jointly. Second, the silhouette language branch primarily acts as a feature enhancer, leveraging sparse and discrete cues embedded from LLMs, which complement the continuous features extracted by the visual branch. Moreover, Freezing the LLM preserves its pretrained representational capacity, while training the gait backbone facilitates better alignment between visual and text embedding space.

\textbf{Size of the Silhouette Vocabulary.} Results in Table~\ref{table_pixels_ablation_study} indicate that a larger silhouette vocabulary further improves performance. This improvement can be attributed to the increased embedding capacity afforded by a larger vocabulary, which expands the vector space to capture more combinations of tokens representing motion patterns.

\textbf{Scaling Behavior of the Silhouette Language Model.} As shown in Table~\ref{table_llm_model_ablation_study}, the results demonstrate the impact of LLM size on performance, revealing that larger models consistently yield higher performance. Specifically, Qwen3-Embedding with 0.6B and 4B parameters set embedding dimensions to 1024 and 2560, respectively, whereas each gait silhouette contains 2816 pixels. Results indicate that a higher dimension can reduce information loss during the compression from tokens to the LLM embedding space, thereby enabling richer and more discriminative representations that further improve performance.
\subsection{Visualization}  
To further validate the effectiveness of SilLang, we visualize the token similarity between silhouettes and natural language, as well as the changes in embedding distance before and after incorporating the silhouette language branch.

\begin{figure}[htb]
	\centering
	\includegraphics[width=\linewidth]{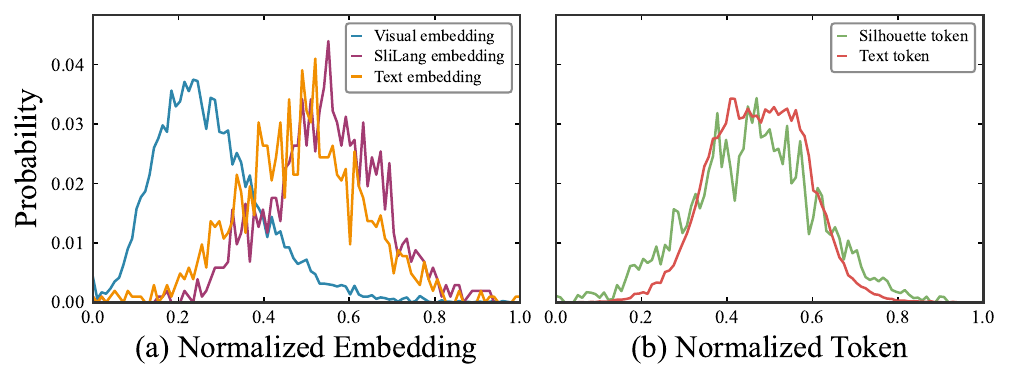}
	\caption{Distribution of normalized embeddings and tokens across different branches. (a) Embeddings in visual, silhouette language and text branches; (b) Input tokens for LLMs.}
	\label{figure_ev_token_space}
\end{figure}

\textbf{Token Similarity of Silhouette and Language.} As illustrated in Figure~\ref{figure_ev_token_space}, we compare the normalized distributions of the visual embeddings ($\boldsymbol{emb}_v$), silhouette language embeddings ($\boldsymbol{emb}_t$), text embeddings, encoded silhouette tokens ($\boldsymbol{token}_{sil}$) and text tokens. $\boldsymbol{emb}_v$, $\boldsymbol{emb}_t$ and $\boldsymbol{token}_{sil}$ are extracted from silhouettes in the Gait3D~\cite{ref_gait3d} dataset, while the text embeddings and tokens are obtained by encoding natural language text through Qwen3-Embedding-0.6B~\cite{ref_qwen3_embedding}.

In Figure~\ref{figure_ev_token_space} (a), $\boldsymbol{emb}_t$ obtained from the language model exhibit a distribution closely aligned with that of text embeddings, while both differ substantially from the visual embeddings $\boldsymbol{emb}_v$.
This observation indicates that SilLang preserves the embedding capability of the LLM.
In Figure~\ref{figure_ev_token_space} (b), the encoded silhouettes tokens $\boldsymbol{token}_{sil}$ show a distribution highly similar to that of text tokens, confirming the similarity between silhouettes and language.
This suggests that silhouettes can be encoded as a structure-driven language.
Moreover, the results validate that the $EA$ module successfully constructs an effective silhouette vocabulary.
\begin{figure}[htb]
	\centering
	\includegraphics[width=\linewidth]{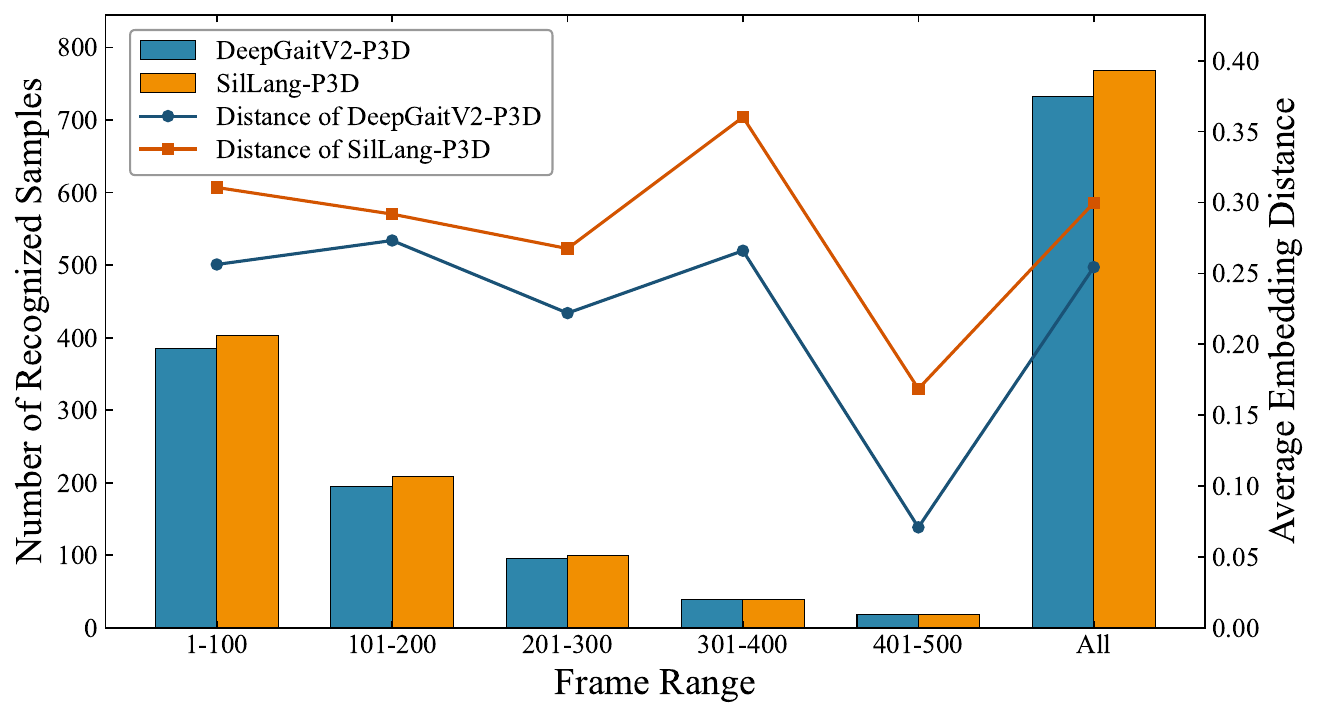}
	\caption{Performance improvement of SilLang on Gait3D. The number of correctly recognized samples in test set are partitioned according to the frame-length ranges. And the average distance is defined as the mean distance between negative samples minus that between positive samples within each frame-length range.}
	\label{figure_frame_distance}
\end{figure}
\textbf{Analysis of the Improving Accuracy.} As shown in Figure~\ref{figure_frame_distance}, SilLang substantially improves Rank-1 accuracy for short sequences (1–100 frames). This improvement arises because the training process employs a fixed input length of 30 frames, indicating that the $Adapter$ in silhouette language branch effectively adapts the $\boldsymbol{emb_t}$ tailored to a specific sentence length. Moreover, incorporating the SilLang increases the distance between positive and negative sample embeddings, thereby improving class separability and enhancing recognition accuracy.
\subsection{Discussion}
Our findings suggest that sparse visual data and natural language share fundamental commonalities in encoding complexity and representation discreteness within their encoding spaces. Similar to DeepSeek-OCR~\cite{ref_deepseekOCR}, which demonstrates that visual models can be extended to text understanding, our work shows that language models can, conversely, embed sparsely encoded binary gait silhouette images. This structural symmetry implies that both modalities may operate within comparable basis vector spaces, enabling the unification of visual sparsity and linguistic discreteness. 
Furthermore, ESM3~\cite{ref_esm3} and Evo~\cite{ref_evo} extract discrete tokens from proteins and DNA based on amino-acid and nucleotide sequences, supporting that discrete data can be encoded within LLM frameworks.
Such a perspective not only explains the effectiveness of proposed SilLang but also points to new opportunities for integrating sparse visual and linguistic representations within a shared encoding framework.

%% file: sec/5_conclusion.tex
\section{Conclusion}
\label{sec:conclusion}
\fussy
Gait silhouettes are represented as binary images and can be encoded into binary gait codes, sharing a discrete and sparse vector space similar to word vocabularies.
This structural similarity enables LLMs to enhance continuous visual gait features using sparse text-driven embeddings, while allowing gait recognition models to capture fine-grained temporal motion variations induced by subtle contour and velocity shifts across frames.
However, the token density and frequency distributions differ between silhouette and text tokens.
To address this, we propose the Contour-Velocity Tokenizer that adjusts these distributions and effectively constructs an implicit silhouette vocabulary.
Experimental results demonstrate that binary gait silhouettes can be translated into language, thereby enhancing gait representations and the recognition performance.

%% file: sec/X_suppl.tex
\clearpage
\maketitlesupplementary
\section{Additional Visualizations}
\label{sec:rationale}
\sloppy
\subsection{Token Density and Frequency}
\begin{figure}[htb]
	\centering
	\includegraphics[width=\linewidth]{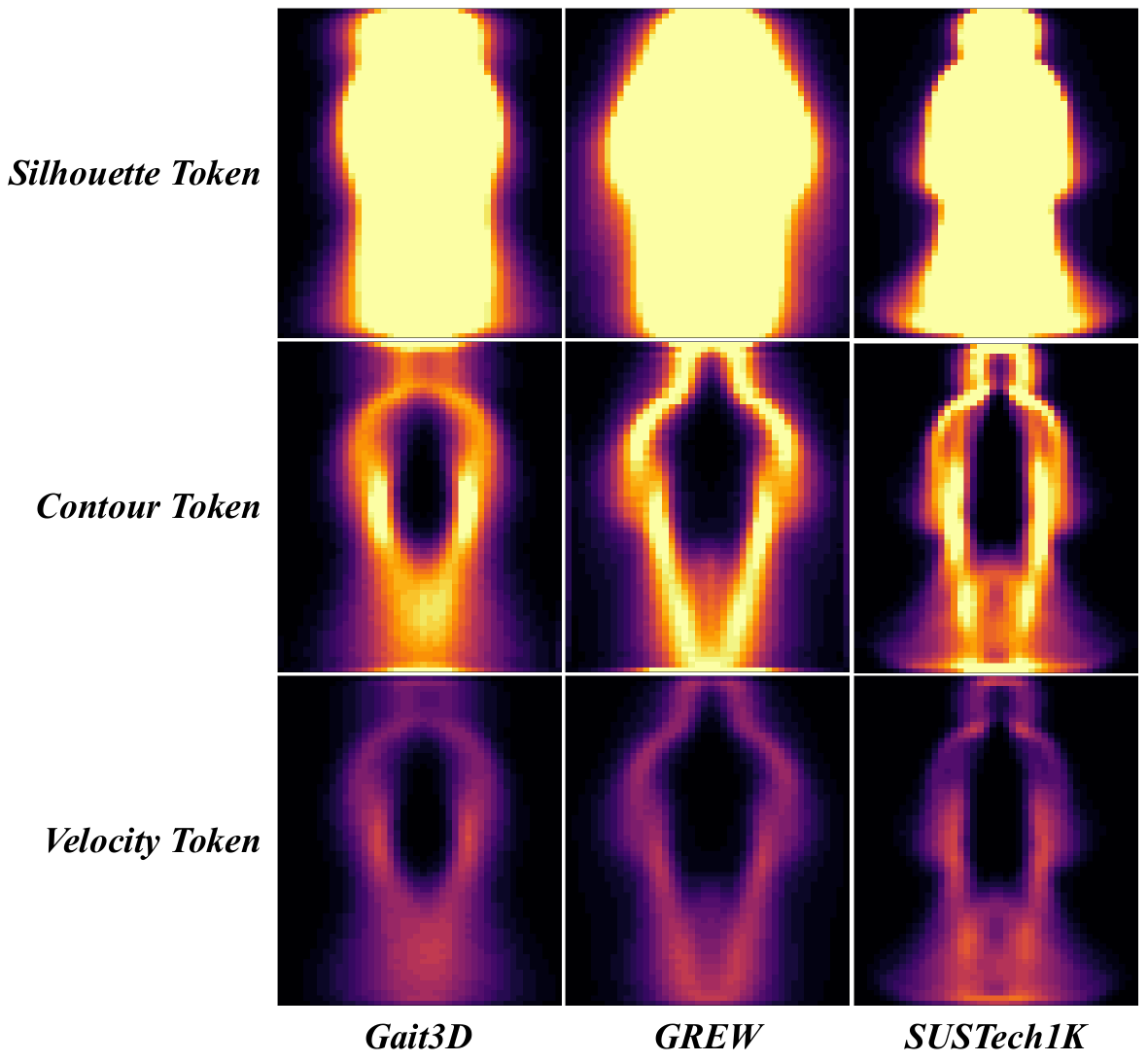}
\caption{Normalized frequency heatmaps of $\boldsymbol{s}(t)$, $\boldsymbol{c}(t)$, and $\boldsymbol{v}(t)$ on Gait3D~\cite{ref_gait3d}, GREW~\cite{ref_grew}, and SUSTech1K~\cite{ref_sustech1k}.}
	\label{figure_scv_token_frequency_distribution}
\end{figure}

The frequency heatmaps of silhouette tokens $\boldsymbol{s}(t)$, contour tokens $\boldsymbol{c}(t)$, and velocity tokens $\boldsymbol{v}(t)$ across Gait3D~\cite{ref_gait3d}, GREW~\cite{ref_grew}, and SUSTech1K~\cite{ref_sustech1k} are presented in Figure~\ref{figure_scv_token_frequency_distribution}.
In these heatmaps, the area of the high-intensity regions indicates the token density, whereas the intensity within these regions represents the token frequency.
To maintain a consistent brightness range across token types, the heatmaps are normalized according to the frequency range of $\boldsymbol{c}(t)$, as $\boldsymbol{c}(t)$ exhibits an intermediate distribution between the much higher frequencies of $\boldsymbol{s}(t)$ and the substantially lower frequencies of $\boldsymbol{v}(t)$.
In the normalized heatmaps, the three token types exhibit a consistent pattern across all datasets, demonstrating the robustness of the proposed CVT tokenizer.
The silhouette token density $P_s(1)$ within each frame is exceedingly high, far surpassing the contour token density $P_c(1)$, which itself remains considerably higher than the velocity token density $P_v(1)$.
Spatially, contour tokens concentrate along human-body boundaries, forming a distinct contour band, while velocity tokens exhibit a similar boundary-aligned pattern but with much lower frequency.
These observations indicate that the proposed CVT tokenizer reduces the excessive density of silhouette tokens and enhances low-to-mid frequency signals, enabling the capture of subtle contour and motion shifts that silhouette tokens alone can hardly encode.
\begin{figure}[tb]
	\centering
	\includegraphics[width=\linewidth]{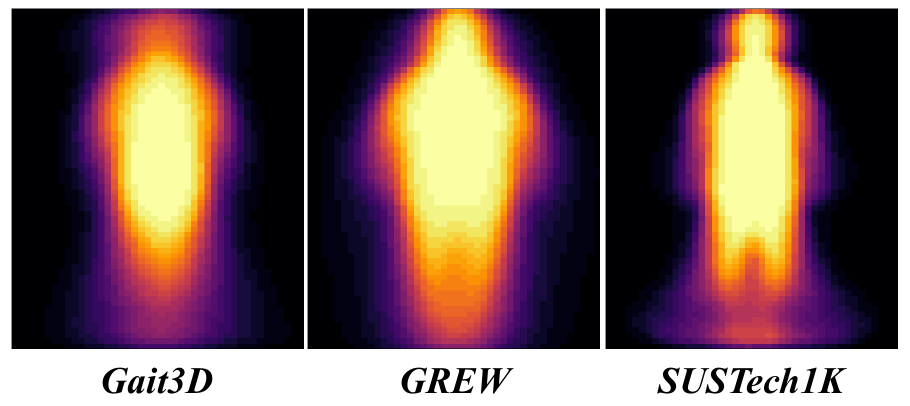}
	\caption{Original frequency heatmaps of $\boldsymbol{s}(t)$ on Gait3D~\cite{ref_gait3d}, GREW~\cite{ref_grew} and SUSTech1K~\cite{ref_sustech1k}.}
	\label{figure_sil_token_frequency_distribution}
\end{figure}

As shown in Figure~\ref{figure_sil_token_frequency_distribution}, the original frequency heatmaps exhibit slight variations in token distributions across datasets, indicating that the statistical properties of silhouette, contour, and velocity tokens are dataset dependent.
Future work may therefore explore the design of a cross-dataset adaptive tokenizer that first distills dataset-invariant gait characteristics and then adjusts token distributions to match different datasets.
Another promising direction is to jointly train the silhouette tokenizer on multiple datasets, which would enhance its generalization ability and mitigate dataset-induced distribution bias.
In addition, future research may explore data augmentation strategies that make fuller use of the black (inactive) regions in silhouettes to expand the diversity of token patterns within the silhouette vocabulary.
Such strategies would not only reduce the overly high frequencies of silhouette tokens but also activate the low-frequency regions, enabling more complete use of the entire token space.
These efforts may facilitate a more balanced frequency distribution and promote a closer alignment with the statistical properties of natural language.
\subsection{Silhouette Vocabulary}
\begin{figure*}[tb]
	\centering
	\includegraphics[width=\linewidth]{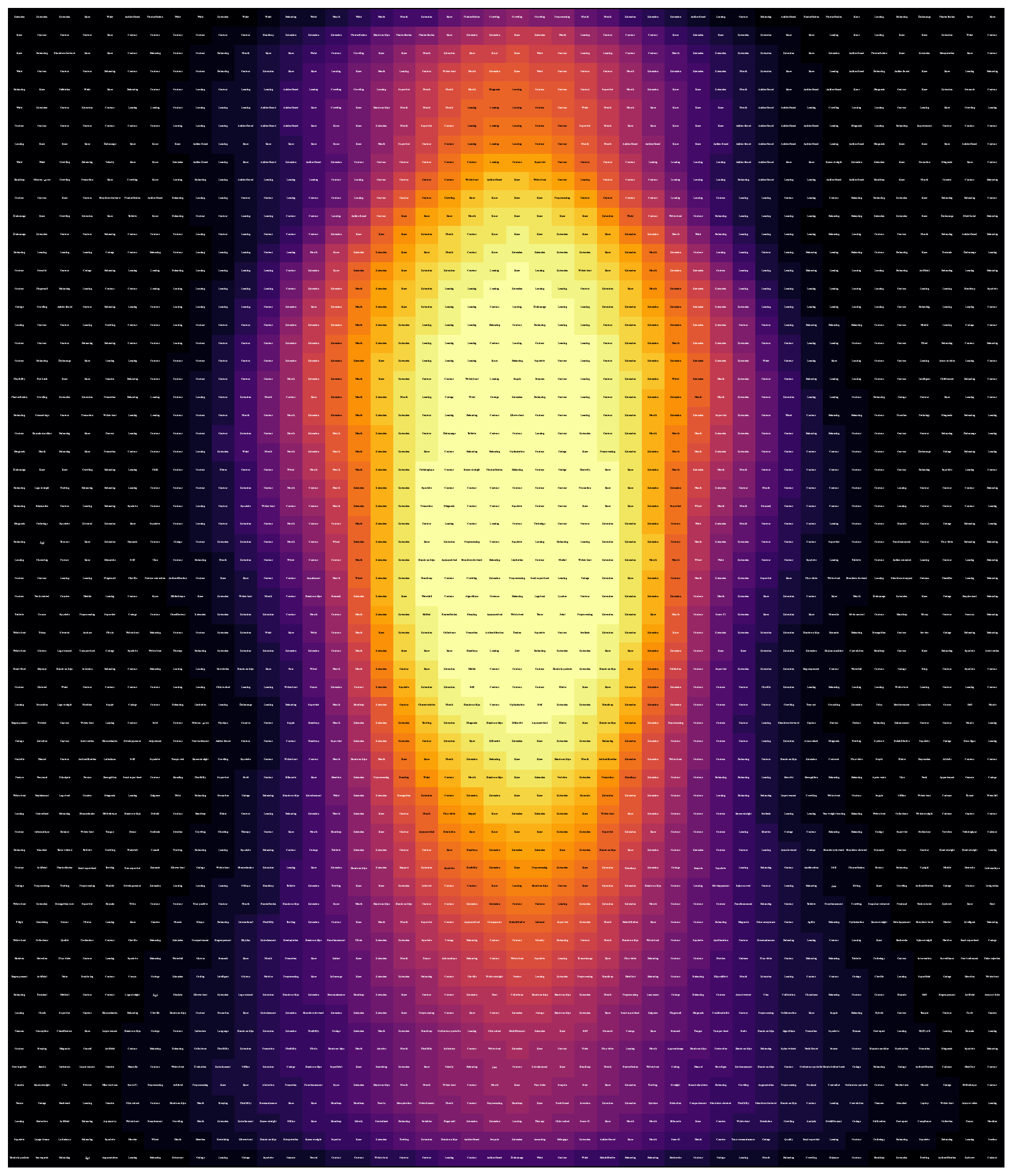}
	\caption{Visualization of the silhouette vocabulary ($64\times 44$). The frequency heatmap is extracted from Gait3D~\cite{ref_gait3d}. These text tokens are not directly provided as encoded silhouette tokens to the LLMs, which are measured by embedding similarity of tokens for visualization.
    }
	\label{figure_visualized_vocabulary}
\end{figure*}

As shown in Figure~\ref{figure_visualized_vocabulary}, the silhouette vocabulary is constructed by measuring the embedding similarity between encoded silhouette tokens and thousands of text tokens.
Since the encoded silhouette tokens are not solely used to represent silhouette, their composition implicitly encodes information about the individual and motion pattern, which can be interpreted as a compound word.
Moreover, within this token embedding space, multiple encoded silhouette tokens can correspond to the same text token, allowing similar tokens to represent different compound words, which is consistent with the properties of the silhouette vocabulary described above.
